\documentclass[11pt]{article}

\usepackage[final]{acl}

\usepackage{times}
\usepackage{latexsym}

\usepackage[T1]{fontenc}

\usepackage{mdwlist}

\usepackage[utf8]{inputenc}

\usepackage{microtype}

\usepackage{inconsolata}
\usepackage{enumitem}

\usepackage{graphicx}

\usepackage{booktabs,tabularx,threeparttable,rotating,adjustbox}
\newcolumntype{L}{>{\raggedright\arraybackslash}X}

\usepackage{array}
\usepackage{ragged2e}
\newcolumntype{P}[1]{>{\RaggedRight\arraybackslash}p{#1}}

\usepackage{siunitx}
\usepackage{amsmath}

\newcommand{\regMTML}{\textsc{MT–ML}} 
\newcommand{\regMTXL}{\textsc{MT–CL}} 
\newcommand{\regCTML}{\textsc{CT–ML}} 
\newcommand{\regXTXL}{\textsc{CT–CL}} 
\usepackage{subcaption}

\usepackage[ruled,vlined,linesnumbered]{algorithm2e}

\title{Donors and Recipients: On Asymmetric Transfer Across Tasks and Languages with Parameter-Efficient Fine-Tuning}

\author{
\textbf{Kajetan Dymkiewicz\textsuperscript{1}} \quad
\textbf{Ivan Vuli\'c\textsuperscript{1}} \quad
\textbf{Helen Yannakoudakis\textsuperscript{2}} \\
\textbf{Eilam Shapira\textsuperscript{3}} \quad
\textbf{Roi Reichart\textsuperscript{3}} \quad
\textbf{Anna Korhonen\textsuperscript{1}} \\
\textsuperscript{1}University of Cambridge, Cambridge, UK \quad
\textsuperscript{2}King's College London, London, UK
\\
\textsuperscript{3}Technion--Israel Institute of Technology, Haifa, Israel
\\
{\ttfamily\small
ktd27@cam.ac.uk \quad
iv250@cam.ac.uk \quad
helen.yannakoudakis@kcl.ac.uk
}
\\
{\ttfamily\small
eilamshapira@campus.technion.ac.il \quad
roiri@technion.ac.il \quad
alk23@cam.ac.uk
}
}

\begin{document}
\maketitle

\begin{abstract}
Large language models (LLMs) perform strongly across tasks and languages, yet how improvements in one task or language affect other tasks and languages remains poorly understood. We conduct a controlled LoRA fine-tuning study across multiple open-weight LLM families and scales, using a standardised grid of 11 languages and four benchmarks. We fine-tune each model on a single task--language \emph{source}, then evaluate it on all other task--language \emph{target} pairs to measure transfer.
We decompose transfer into three regimes: (i) Matched-Task (Cross-Language), (ii) Cross-Task (Matched-Language), and (iii) Cross-Task (Cross-Language). Single-source fine-tuning yields a net positive uplift across regimes, but the gains are strongly asymmetric. Matched-Task (Cross-Language) transfer emerges as the most effective and structurally regular regime, with transfer magnitude driven principally by the identity of the target language rather than model architecture. We identify a stable coarse-grained hierarchy in which some task and language targets consistently absorb gains from diverse sources, while others remain relatively isolated. These results imply that effective fine-tuning requires accounting for donor--recipient roles to maximise downstream gains while limiting collateral degradation in other capabilities.
\end{abstract}

\section{Introduction}
\label{sec:intro}
Large language models (LLMs) have become a cornerstone of modern AI, exhibiting impressive capabilities across a wide range of tasks \citep{openai2024gpt4technicalreport,NEURIPS2020_1457c0d6}.
In parallel, parameter-efficient fine-tuning (PEFT) methods such as LoRA \citep{peft,hu2022lora} provide an effective way to specialise models. However, how updates from a particular source propagate to other tasks and languages remains under-explored. Evidence from related settings suggests that such broader effects may not be uniformly beneficial. Instruction tuning on multiple tasks can reduce subsequent transfer to unseen tasks \citep{mueller-etal-2024-multi}, while sequential adaptation can cause catastrophic forgetting of previously acquired capabilities \citep{goodfellow2014empiricalinvestigationcatastrophicforgetting}. These risks are especially salient in multilingual models, whose performance can already vary substantially across languages \citep{pmlr-v119-hu20b,hu-etal-2025-quantifying}.
Multilingual pre-training approaches such as mBART \citep{liu-etal-2020-multilingual-denoising} and XLM-R \citep{conneau-etal-2020-unsupervised}, together with joint multilingual and multi-task training, learn from mixtures of languages or tasks. Their aggregate outcomes therefore do not directly isolate the effect of one particular source on each possible target. This motivates a more granular question: after adapting a model on one task--language pair, which other task--language capabilities improve, remain unchanged, or degrade?

\begin{figure*}[t]
  \centering
  \includegraphics[width=\textwidth]{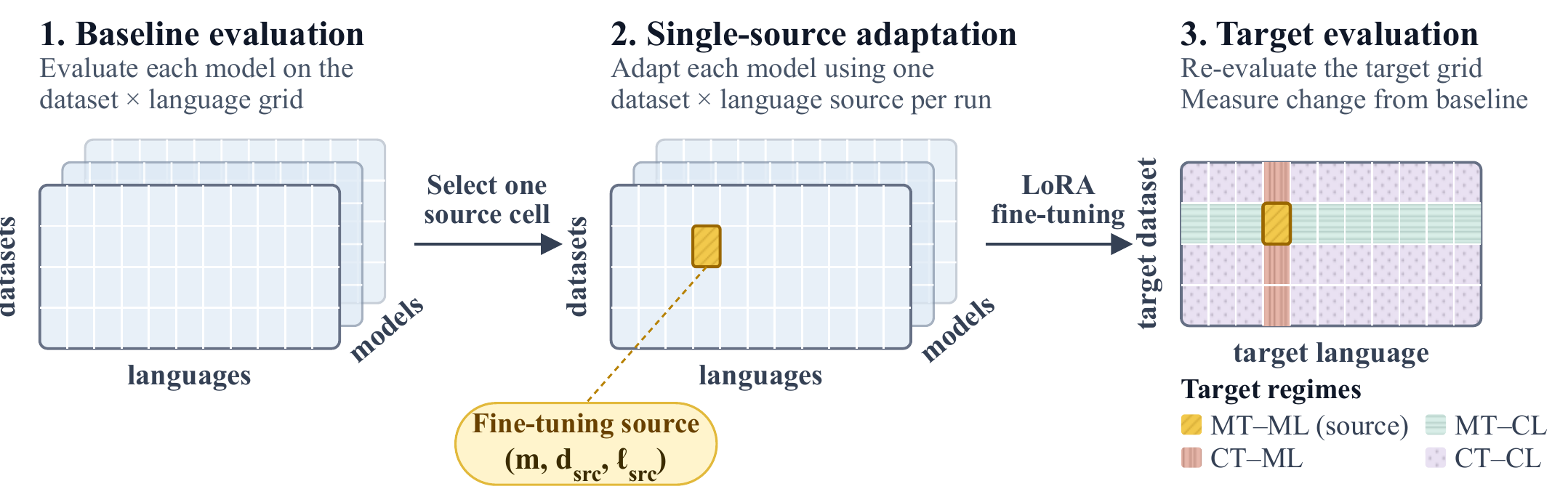}
  \caption{Overview of the experimental pipeline. (1) We first evaluate each instruction-tuned model across the dataset--language grid to establish pre-adaptation baselines. (2) For each model, we then select one dataset--language source $(d_{\mathrm{src}}, \ell_{\mathrm{src}})$ and fine-tune with LoRA. (3) We re-evaluate the adapted model across the target grid and measure percentage-point changes from the corresponding baselines. The gold \regMTML{} cell marks the on-source evaluation; the other colours and patterns identify the three transfer regimes. Repeating this process for every source cell and model yields ordered percentage-point changes for the regime and donor--recipient analyses.}
  \label{fig:experimental-pipeline}
\end{figure*}

In order to address this gap, we study a task--language grid of four benchmarks and eleven languages across three model families and multiple scales. Every benchmark is represented in every language, and each task--language pairing (hereafter, a \emph{cell}) serves as both a source and a target. For each model, we hold the starting instruction-tuned checkpoint and adaptation recipe fixed across source cells. In each run, we fine-tune on one source cell and evaluate the resulting model across the target grid. This design yields a directed transfer map, allowing us to assess both the impact of adaptation and the stability of its effects across models and scales (Figure~\ref{fig:experimental-pipeline}).
Specifically, we seek to answer the following research questions:

\paragraph{Impact and Structure of Transfer} How does fine-tuning on one source cell affect performance elsewhere? How do these effects vary across transfer regimes? Which tasks or languages consistently act as strong donors or recipients? When does fine-tuning benefit or harm other task--language pairs, and to what extent?

\paragraph{Stability and Determinants of Transfer Patterns}
How stable are transfer patterns across model families and sizes? How much of the variation in transfer is attributable to the model versus the transfer source and target, and how does this balance shift across transfer regimes?

\paragraph{Contributions}\mbox{}\par

\noindent\textbf{Transfer impact.}
We find a structured yet modest global uplift from fine-tuning and a pronounced asymmetry between same-task, cross-language and off-task regimes. Same-task, cross-language transfer is positive on average, with high positive-transfer rates and substantially lower harm rates than cross-task transfer, which yields smaller gains and markedly higher harm rates.

\noindent\textbf{Donor--recipient structure.}
We reveal systematic donor--recipient structure across tasks and languages: donor strength and recipient receptivity are not interchangeable, and target-side properties dominate variance in transfer strength.

\noindent\textbf{Stability and practical guidance.}
Using a mixed-effects variance decomposition and a rank-based Consistency Index, we characterise the cross-model stability of these patterns.

Taken together, our findings on transfer impact, donor--recipient structure, and stability yield risk-aware fine-tuning heuristics that prioritise matched-task sources, identify fragile recipients, and flag regimes likely to require additional safeguards against collateral harm.

\paragraph{Resources.}
We make our training, evaluation, and analysis code, per-cell transfer results, and benchmark data splits publicly available at \url{https://github.com/kdymkiewicz/Donors-and-Recipients}. For each benchmark, the disjoint training and evaluation splits use the same underlying examples in all eleven languages. The code is released under the Apache License 2.0.

\section{Related Work}
Prior work on knowledge transfer between sources and targets has examined how model adaptation reshapes performance across languages and tasks.

\subsection{Cross-Lingual Transfer}
Multilingual pre-training learns shared representations from mixtures of languages before downstream adaptation. mBART uses multilingual denoising pre-training for sequence-to-sequence transfer \citep{liu-etal-2020-multilingual-denoising}, while XLM-R scales masked language-model pre-training across many languages \citep{conneau-etal-2020-unsupervised}.

These approaches demonstrate the benefits of joint multilingual exposure.
However, research on downstream cross-lingual transfer shows that pre-training distributions and adaptation choices can produce uneven outcomes across languages. \citet{malkin-etal-2022-balanced} show that pre-training languages can act as asymmetric donors in zero-shot transfer, while \citet{chua2025crosslingualcapabilitiesknowledgebarriers} identify a ``cross-lingual knowledge barrier'': models show surface-level cross-lingual abilities but struggle with deeper cross-lingual knowledge transfer. Zero-shot instruction-tuning work \citep{chirkova-nikoulina-2024-zero-shot} shows that English-only tuning can support cross-lingual generalisation, but target-language response quality remains uneven. \citet{aggarwal2025languagemodelsfactualitydepends} further show that factual accuracy itself is language-dependent, so cross-lingual generalisation does not guarantee comparable reliability across languages.

Collectively, these studies show that cross-lingual transfer varies systematically with pre-training distributions, adaptation choices, and target-language properties, but leave unresolved how these language-level asymmetries interact with transfer across tasks.

\subsection{Task Transfer and Parameter Efficiency}
Beyond cross-linguality, cross-task transfer introduces its own trade-offs. Multi-task instruction tuning can improve generalisation to unseen tasks \citep{wei2022finetuned}, but fine-tuning on a new task can erode previously acquired capabilities through catastrophic forgetting \citep{li-etal-2024-revisiting}.

Parameter-efficient fine-tuning (PEFT) methods such as LoRA \citep{hu2022lora} freeze pretrained weights and represent selected updates with trainable low-rank factors, while modular approaches such as AdapterFusion learn to compose multiple task adapters without destructive overwriting \citep{pfeiffer-etal-2021-adapterfusion}. Yet, gains in average performance need not be uniformly beneficial: for example, \citet{zhang2024exploringaccuracyfairnesstradeofflarge} find that optimising solely for accuracy can degrade fairness.

Multi-task and modular approaches typically evaluate aggregate generalisation or retention across tasks. We instead isolate each task--language source and compare all ordered source--target pairs, exposing directional asymmetries and donor--recipient structure hidden by aggregation.

\section{Methodology}
\label{sec:methodology}
We first describe the model suite and benchmark grid, then detail the experimental procedure and transfer measures.

\subsection{Models and Benchmark Grid}
\paragraph{Model Suite}
To analyse the stability of transfer patterns across architectural designs and scales, we evaluate nine instruction-tuned, open-weight models from three families. The model suite comprises Llama~3.2 (1B and 3B) \citep{meta2024llama32modelcard} and Llama~3.1 (8B) \citep{grattafiori2024llama3herdmodels}; Qwen~2.5 (0.5B, 1.5B, 3B, and 7B) \citep{qwen2025qwen25technicalreport}; and Gemma~3 (1B and 4B) \citep{gemmateam2025gemma3technicalreport}.

\paragraph{Benchmark Grid}
We use four parallel multilingual benchmarks, each covering the same 11 languages. These are English (en), German (de), Spanish (es), French (fr), Italian (it), Portuguese (pt), Indonesian (id), Chinese (zh), Bengali (bn), Hindi (hi), and Arabic (ar).

The four benchmarks are \textit{ARC-Challenge} \citep{clark2018thinksolvedquestionanswering} (reasoning); \textit{HellaSwag} \citep{zellers-etal-2019-hellaswag} (commonsense inference); \textit{TruthfulQA} \citep{lin-etal-2022-truthfulqa} (truthfulness and hallucination); and \textit{Global-MMLU-Lite} \citep{singh-etal-2025-global} (multidisciplinary knowledge). We refer to \textit{Global-MMLU-Lite} as \textit{Global-MMLU} throughout.
For \textit{ARC-Challenge} and \textit{HellaSwag}, we use the multilingual machine-translated variants released with Okapi \citep{lai-etal-2023-okapi}.
For \textit{TruthfulQA}, we use the separately released multilingual variant.\footnote{
\url{https://huggingface.co/datasets/alexandrainst/m_truthfulqa}}

\paragraph{Dataset Alignment}
To ensure comparable transfer directions, we construct custom, parallel training and evaluation splits with strict example-level alignment across languages. For each benchmark, we select disjoint sets of example IDs for training and evaluation and use the same two sets in every language. This prevents an example used for training in one language from appearing in evaluation in another. We also approximately match the training budget across task--language pairings (see Table~\ref{tab:benchmarks-languages} for dataset characteristics). Appendix~\ref{app:dataset_construction} provides the alignment procedure and the released split pools used for each benchmark (Algorithm~\ref{alg:alignment}, Table~\ref{tab:split_sources}).

\subsection{Experimental Procedure}
Figure~\ref{fig:experimental-pipeline} summarises the three-stage procedure detailed below; implementation details are provided in Appendix~\ref{app:impl-details}.

\paragraph{Baseline Evaluation}
First, we evaluate each starting instruction-tuned model on every cell in the task--language grid. These scores form the pre-adaptation baselines against which we measure subsequent changes.

\paragraph{Single-Source Adaptation}
For each model--source-cell configuration, we run three experiment seeds with a fixed train--validation split and average each ordered source--target delta across seeds. We fine-tune for up to three epochs using a fixed adapter configuration and optimisation settings to isolate source-selection effects (see Appendix~\ref{app:lora_rank}). We retain the lowest-validation-loss checkpoint for transfer evaluation.

  \paragraph{Target Evaluation}
  We evaluate each adapted model on the full task--language grid, including its source cell, using the same protocol as during baseline evaluation. Comparing the resulting scores with their pre-adaptation counterparts yields the ordered source--target effects analysed below.

\subsection{Transfer Measures}
\label{sec:evaluation}

\paragraph{Performance Metrics}
For an instruction-tuned model $m$, let $m_{\text{ft}}$ denote its variant fine-tuned on a single source cell $(d_{\text{src}}, \ell_{\text{src}})$. We measure transfer as the absolute percentage-point change:
\[
\Delta = 100 \cdot [s(m_{\text{ft}}) - s(m)],
\]
where $s(\cdot)$ is the score. We also report two complementary summary measures:
\begin{itemize}[leftmargin=*]
    \item \textbf{Positive-transfer rate:} percentage of target cells with $\Delta > 0$;
    \item \textbf{Harm rate:} percentage of target cells with $\Delta \leq -1.0$~pp.
\end{itemize}

\paragraph{Transfer Regimes}
For the transfer analysis, we exclude the source cell and partition the remaining target cells $(d, \ell)$ into three regimes:
\begin{itemize}[leftmargin=*]
    \item \textbf{Matched-Task / Cross-Language (\regMTXL{}):} same dataset, different language ($d = d_{\text{src}}$, $\ell \neq \ell_{\text{src}}$);
    \item \textbf{Cross-Task / Matched-Language (\regCTML{}):} different dataset, same language ($d \neq d_{\text{src}}$, $\ell = \ell_{\text{src}}$);
    \item \textbf{Cross-Task / Cross-Language (\regXTXL{}):} different dataset and different language ($d \neq d_{\text{src}}$, $\ell \neq \ell_{\text{src}}$).
\end{itemize}
The direct matched-task, matched-language case $(d = d_{\text{src}}, \ell = \ell_{\text{src}})$ forms a fourth regime (\regMTML{}). We exclude it from the transfer computations and treat it as the on-source baseline.
We use the task-first abbreviations \regMTML{}, \regMTXL{}, \regCTML{}, and \regXTXL{} throughout, including in tables, and refer to \regCTML{} and \regXTXL{} collectively as the \emph{off-task} regimes.

\paragraph{Donor--Recipient Scores}
Within the \regMTXL{} regime, we summarise each language's outgoing and incoming transfer using two scores. The \textbf{Language Donor Score} is the mean seed-averaged $\Delta$ over cases in which the language serves as the source, whereas the \textbf{Language Recipient Score} is the corresponding mean over cases in which it serves as the target. We aggregate both scores across datasets, model families, and scales. Task-level donor and recipient scores are defined analogously within the \regCTML{} regime, holding the language fixed.

\begin{figure*}[t!]
  \centering
  \includegraphics[width=\textwidth]{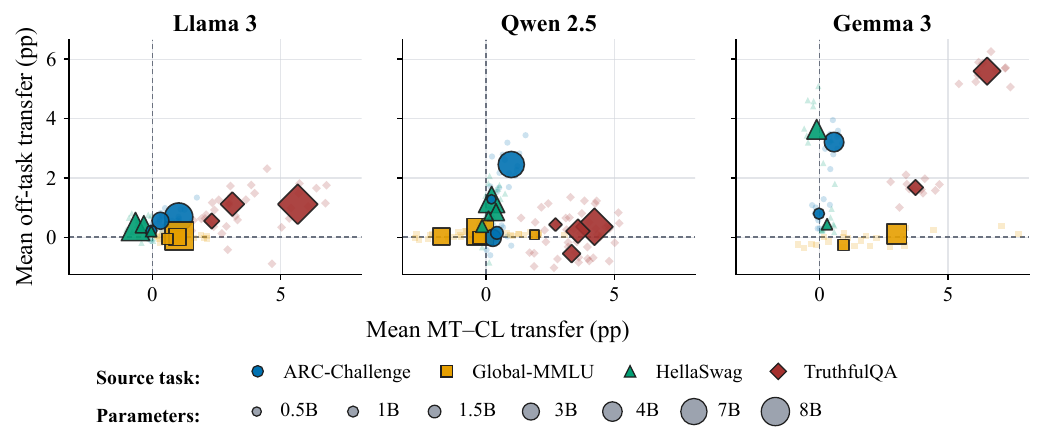}
  \caption{Matched-task and off-task transfer profiles by model family. Faint markers show individual model--source-cell configurations, aggregated over the three experiment seeds; outlined markers average the source languages for each model and source task. Colour and shape identify the source task, and marker size denotes model scale. \textit{Global-MMLU} remains near zero off-task transfer, whereas the other tasks export gains to off-task recipients in strongly model-dependent magnitudes.}
  \label{fig:pareto_onoff}
\end{figure*}

\section{Results and Analysis}
We first quantify aggregate transfer outcomes (Section~\ref{sec:impact}), then examine the directed donor--recipient structure (Section~\ref{sec:donor-structure}), and finally assess its stability and robustness across models, dataset construction, and adaptation settings (Section~\ref{sec:stability}).

\subsection{Aggregate Transfer Outcomes}
\label{sec:impact}
We begin with the aggregate effect across all models, source cells, and target task--language cells.
Overall, single-source fine-tuning yields a modest but clearly positive mean uplift of $+0.89$~pp, with a positive-transfer rate of $59.7\%$. This indicates that, on average, single-source adaptation preserves or slightly enhances off-target performance.

\paragraph{Transfer Regimes}
Table~\ref{tab:bucket-global} quantifies the performance across the three transfer regimes.
\regMTXL{} is the most reliable source of improvement, combining the largest average gain with the highest positive-transfer rate and lowest harm rate.
By contrast, the two cross-task regimes remain net positive but show smaller gains, lower positive-transfer rates, and greater harm.
Notably, \regMTXL{} recovers roughly 62\% of the gain available from direct \regMTML{} LoRA fine-tuning.

\begin{table}[h!]
\centering
{\small
\setlength{\tabcolsep}{4pt}%
\renewcommand{\arraystretch}{1.15}%
\begin{tabular}{lcccc}
\toprule
\textbf{Regime} & \textbf{Mean $\Delta$} & \textbf{Median $\Delta$} & \textbf{Positive \%} & \textbf{Harm \%} \\
\midrule
\regMTXL & +1.25 & +0.50 & 66.40 &  7.10 \\
\regCTML & +0.81 & +0.17 & 56.90 & 11.40 \\
\regXTXL & +0.78 & +0.17 & 57.70 &  9.50 \\
\bottomrule
\end{tabular}
}
\caption{Aggregate transfer outcomes by regime. \regMTXL{} combines the largest gains with the highest positive-transfer rate and lowest harm rate; both off-task regimes remain net positive but yield smaller gains.}
\label{tab:bucket-global}
\end{table}

\paragraph{Matched-Task and Off-Task Transfer Profiles}
For each model--source-cell configuration, after averaging deltas over the three experiment seeds, we compute (i) mean \regMTXL{} transfer across the other target languages and (ii) mean off-task transfer across all cross-task targets. Positive values indicate beneficial transfer, while negative values indicate performance degradation.
Figure~\ref{fig:pareto_onoff} organises these profiles by model family and foregrounds the model--task means. The complete configuration-level encoding is shown in Figure~\ref{fig:pareto-onoff-full} in the appendix, while exact per-cell values are available in the released result artefact described in Appendix~\ref{app:licenses}.

The source task strongly structures the resulting profiles. \textit{TruthfulQA} produces the largest \regMTXL{} gains across families, but its off-task transfer varies by model: it is modest or occasionally negative for Qwen~2.5 and large for Gemma~3 4B. \textit{ARC-Challenge} and \textit{HellaSwag} often yield positive off-task transfer despite limited \regMTXL{} gains, particularly for Gemma~3. \textit{Global-MMLU}, by contrast, remains close to zero off-task transfer even when it transfers cross-lingually.

Transfer magnitude also tends to increase with scale, most clearly for Llama~3 and Gemma~3, but the pattern is not monotonic for every Qwen~2.5 task. Thus, source tasks exhibit distinct donor profiles shaped by model family and scale, rather than a universal specialisation--generalisation trade-off.

\paragraph{Patterns of Harmful Transfer}
Although \regMTXL{} transfer is helpful on average, 7.1\% of cases incur a performance decline of at least one percentage point and therefore meet our harm criterion.
We decompose these cases to identify fragility patterns. Table~\ref{tab:mtcl_harm_breakdown} provides a fine-grained decomposition across target datasets, target languages, model families, and scale buckets.
This breakdown reveals that harmful outcomes are highly task-dependent. It also shows language-level differences: although all target languages benefit on average, Hindi has a somewhat higher harm rate and lower positive-transfer rate than, for example, Italian, making it a more fragile recipient.
Harm rates also vary mildly with scale: they are highest for medium-sized models, with small and large models slightly more stable.
Finally, stability differs modestly across model families: Llama~3 appears more robust to negative transfer (fewer severe harms), while Gemma~3 and Qwen~2.5 are more sensitive to fine-tuning, achieving different trade-offs between average gains and harmful interference.

\begin{figure}[t!]
  \centering
  \includegraphics[width=\linewidth]{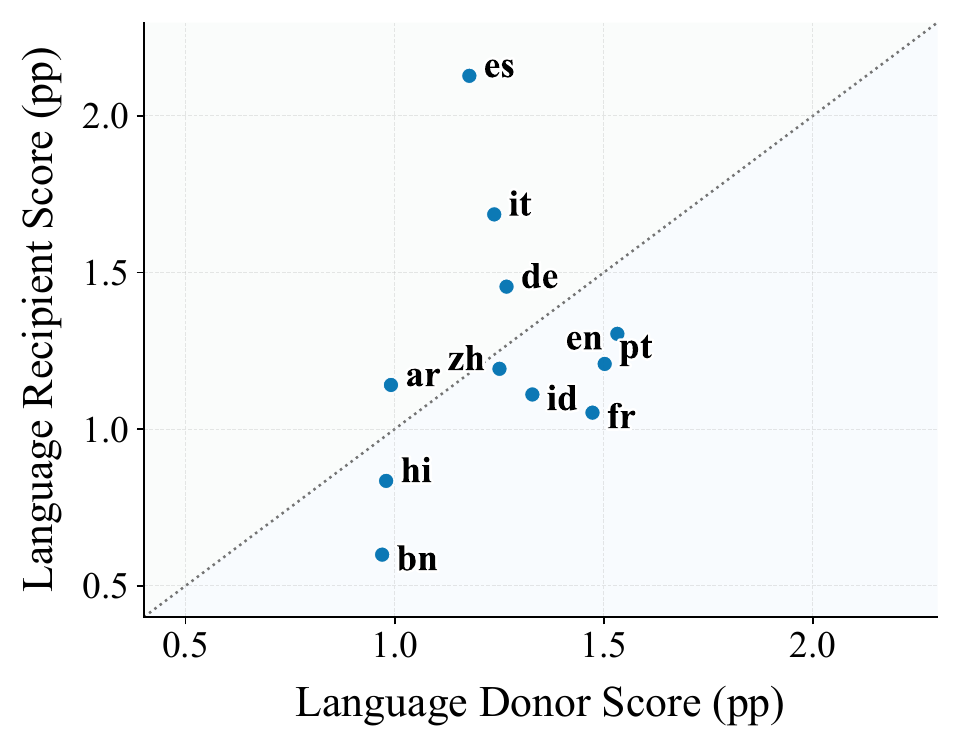}
  \caption{Language donor vs.\ recipient roles in the \regMTXL{} regime. Each language is positioned by its average Donor Score (x-axis) and Recipient Score (y-axis), aggregated across tasks, model families, and scales.
  Spanish and Italian stand out as disproportionately strong recipients relative to their donor strength, while low-resource Indic languages sit well below the diagonal, highlighting a clear hierarchy of cross-lingual roles.}
  \label{fig:lang-roles-scatter}
\end{figure}

\begin{figure*}[t!]
  \centering
  \includegraphics[width=\textwidth]{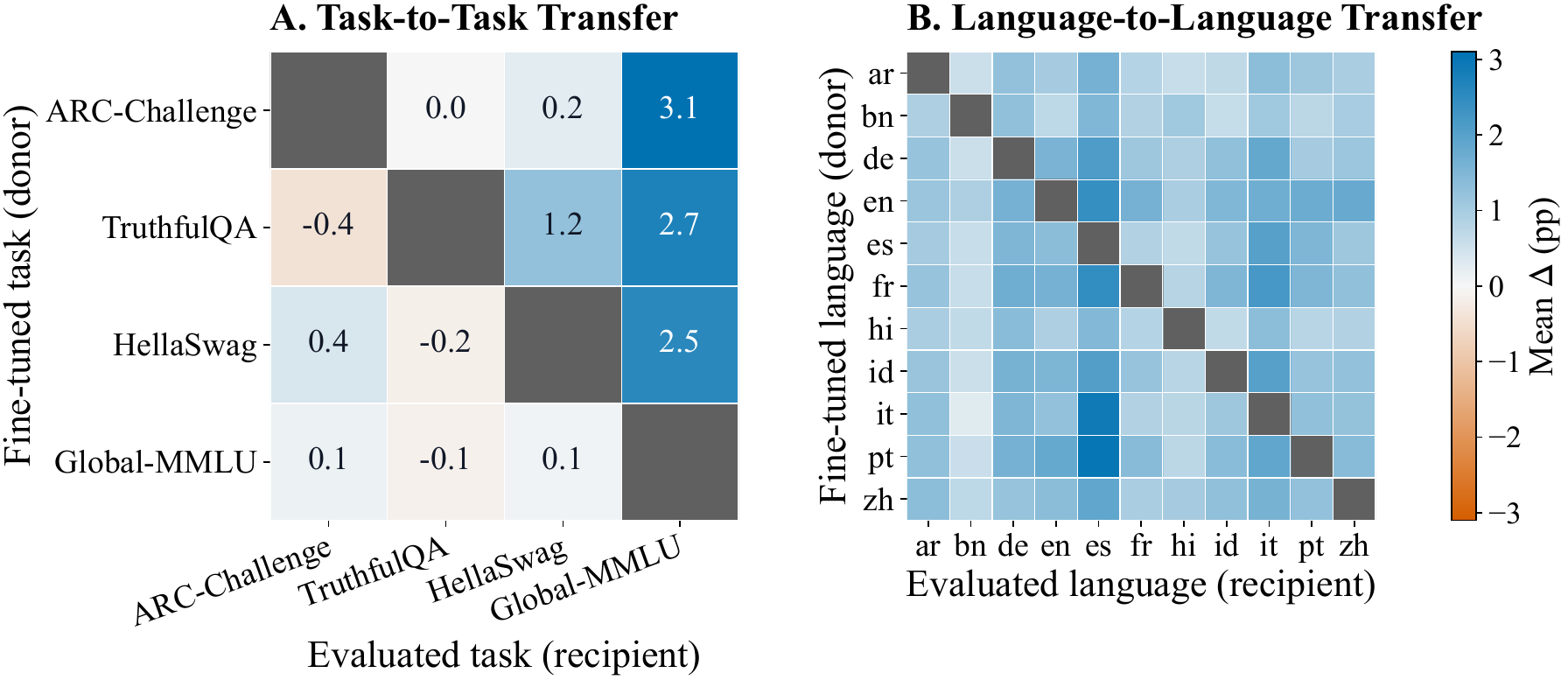}
  \caption{Directed transfer matrices with a shared colour scale. Rows denote fine-tuned donors and columns denote evaluated recipients; cells show mean $\Delta$ in percentage points, with the diagonals masked. Panel A shows task-to-task transfer (\regCTML{}) averaged across languages, model families, and scale buckets; \textit{Global-MMLU} is the strongest task recipient. Panel B shows language-to-language transfer (\regMTXL{}) averaged across datasets, model families, and scale buckets; Spanish and Italian are the strongest language recipients, whereas Bengali and Hindi are the weakest. Exact values appear in Appendix Tables~\ref{tab:task_to_task_transfer} and \ref{tab:lang_to_lang_transfer}.}
  \label{fig:transfer-heatmaps}
\end{figure*}

\subsection{Directed Donor--Recipient Structure}
\label{sec:donor-structure}
The transfer landscape is strongly directional: high donor strength does not imply high recipient receptivity. We analyse this asymmetry first across languages and then across tasks, combining the scores defined in Section~\ref{sec:evaluation} with the pairwise transfer matrices in Figure~\ref{fig:transfer-heatmaps}.

\paragraph{Language-Level Structure}
Figure~\ref{fig:lang-roles-scatter} shows that donor and recipient roles are not interchangeable: although some languages lie near the diagonal, the notable departures reveal substantial differences between outgoing and incoming transfer (see also Table~\ref{tab:lang_donor_recipient}).
Spanish stands out as the strongest recipient, with Italian also scoring highly but at a lower level; both realise large gains from incoming transfer relative to their donor strength, suggesting they are particularly well-positioned to leverage cross-lingual features learnt from other sources.
Other high-resource Western European languages (English, French, German, Portuguese), along with Chinese and Indonesian, form a cluster of strong donors and recipients.
In contrast, Hindi and Bengali lie near the lower end of the recipient axis: while their donor scores are comparable to the rest of the languages, their recipient scores are noticeably lower, indicating that they export gains to other languages while benefiting less from incoming cross-lingual fine-tuning.

In order to examine individual directions, we construct an \regMTXL{} language--language transfer matrix by averaging the seed-averaged deltas for each ordered language pair across tasks, model families, and scale buckets.
Panel~B of Figure~\ref{fig:transfer-heatmaps} is strictly positive but highly stratified. We observe a pronounced ``Romance amplification'' effect: Portuguese and Italian act as exceptionally potent donors for Spanish, yielding the largest gains in the entire study. This suggests that for these high-resource, linguistically close pairs, fine-tuning features are highly transferable.

In contrast, Hindi and Bengali occupy the bottom of the recipient ranking. Bengali is the least receptive language on average: English provides its largest incoming gain, while the remaining donors yield smaller but still positive gains. Although Hindi is phylogenetically closer to Bengali, English transfers more strongly into Bengali than Hindi does, showing that linguistic proximity alone does not determine transfer strength.

\paragraph{Language-Family Aggregation}
With four of the eleven languages, Romance is the largest family represented in our grid. We therefore macro-average the language-level donor and recipient scores by family, treating singleton languages as separate groups (Table~\ref{tab:family-donor-recipient}). Germanic emerges as the strongest donor group, marginally ahead of Romance, whereas Romance is the most receptive. Chinese, Indonesian, and Arabic occupy intermediate positions, while Indic remains comparatively isolated. Thus, the broad core--middle--periphery structure persists after family aggregation, and Romance representation alone does not explain the donor ordering.

\begin{table}[h!]
\centering
\begin{tabular}{lrr}
\toprule
\textbf{Family/group} & \textbf{Donor} & \textbf{Recipient} \\
\midrule
Romance    & 1.35 & 1.52 \\
Germanic   & 1.40 & 1.38 \\
Indic      & 0.97 & 0.72 \\
Chinese    & 1.25 & 1.19 \\
Arabic     & 0.99 & 1.14 \\
Indonesian & 1.33 & 1.11 \\
\bottomrule
\end{tabular}
\caption{Language-family and singleton-group donor and recipient scores in \regMTXL{}. Germanic leads as a donor, Romance as a recipient, and Indic is the weakest recipient.}
\label{tab:family-donor-recipient}
\end{table}

\paragraph{Linguistic Correlates}
\label{sec:ling_drivers}
We further investigate how cross-lingual transfer varies with linguistic distance. Across language pairs, we correlate transfer performance with syntactic, phonological, and inventory distances from the URIEL database \citep{littell-etal-2017-uriel}. As detailed in Appendix~\ref{app:ling_corr}, syntactic distance has the strongest negative association with transfer; phonological distance has a weaker negative association, whereas inventory distance is not statistically significant.

\paragraph{Task-Level Structure}
Donor strength and recipient receptivity also diverge at the task level (Table~\ref{tab:task-donor-recipient}). \textit{TruthfulQA} emerges as the strongest donor but the most fragile recipient, improving other tasks while being vulnerable to negative interference itself, whereas \textit{Global-MMLU} absorbs substantial gains from all other tasks yet contributes almost negligible transfer benefits to them.
This pattern suggests that performance on \textit{Global-MMLU} can benefit from knowledge and reasoning patterns learnt on other tasks. By contrast, the fine-grained response patterns needed to suppress plausible but false completions on \textit{TruthfulQA} appear more brittle and easier to overwrite.
\textit{TruthfulQA}'s donor strength may partly reflect its full-sequence supervision, rather than the answer-token-only supervision used for the other datasets (Appendix~\ref{app:impl-details}).

We summarise these directional effects in the \regCTML{} task--task transfer matrix shown in Panel~A of Figure~\ref{fig:transfer-heatmaps}. Each entry averages the seed-averaged deltas for an ordered task pair across languages, model families, and scale buckets.
The resulting structure is strongly directional. Across all donors, one task (\textit{Global-MMLU}) behaves as a universal beneficiary: it receives consistently large positive gains from every other task but provides almost no benefit in return.
Other tasks act as moderate donors that reliably support this universal beneficiary; they benefit from each other to varying degrees but do not form symmetric pairs in which gains are reciprocated in both directions.

Taken together, these results reinforce the picture from the donor--recipient scores: cross-task interactions form a directed graph, and gains in one direction are not necessarily reciprocated when source and target tasks are reversed.

\subsection{Stability and Robustness}
\label{sec:stability}
We assess stability using a variance decomposition and a rank-based Consistency Index, then test robustness to dataset construction and the adaptation setup.

\paragraph{Variance Decomposition}
To quantify variation in transfer effects, we fit a linear mixed-effects model using restricted maximum likelihood (REML), partitioning variance across model characteristics (family and size), source, and target. Table~\ref{tab:varshares-aggregated} reports each component's share of total $\Delta$ variance.
\begin{table}[h!]
\centering
{\small
\begin{adjustbox}{width=\columnwidth}
\setlength{\tabcolsep}{4pt}%
\begin{tabular}{@{}l
                S[table-format=2.2]
                S[table-format=2.2]
                S[table-format=2.2]
                S[table-format=2.2]@{}}
\toprule
    & \textbf{Overall}
    & \multicolumn{1}{c}{\textbf{\regMTXL}}
    & \multicolumn{1}{c}{\textbf{\regCTML}}
    & \multicolumn{1}{c}{\textbf{\regXTXL}} \\
\midrule
\textbf{Model}    &  8.59 &  4.22 &  8.66 & 10.52 \\
\textbf{Source}   & 10.04 &  8.75 &  2.02 &  5.54 \\
\textbf{Target}   & 40.99 & 75.43 & 48.55 & 57.07 \\
\textbf{Residual} & 40.39 & 11.59 & 40.76 & 26.88 \\
\bottomrule
\end{tabular}
\end{adjustbox}
}
\caption{REML variance shares (\%) of $\Delta$ by model-, source-, and target-level components, overall and by transfer regime. Target-level variation is the largest component in every regime, most strongly in \regMTXL{}.}
\label{tab:varshares-aggregated}
\end{table}

Across all regimes, the target task or language explains the largest variance share, revealing target-side dominance. Model characteristics play only a minor role, suggesting that these patterns reflect the task--language landscape rather than specific architectures.

Target-side dominance is most pronounced in the \regMTXL{} regime, where the target language accounts for most of the variance and the residual component is smaller than in either cross-task regime. This indicates that the receptivity of a target language is the primary determinant of success, far outweighing the choice of donor language or model family within this transfer regime. 

In the cross-task regimes, the target still dominates, but the residual variance increases. This reflects the volatility observed in the task--task transfer heatmap, where specific interactions introduce idiosyncratic noise that is less predictable than linguistic transfer. The source component is negligible in \regCTML{}, reinforcing the finding that while some tasks are better donors than others, the outcome is largely dictated by the target's capacity to absorb transfer in cross-task settings.

\paragraph{Cross-Model Consistency}
Beyond variance shares, we also measure how consistently the nine models
rank recipients for a given donor using a rank-based Consistency
Index (CI; formally defined in Appendix~\ref{app:ci}).
CI values are low across all regimes, indicating that while coarse-grained
tiers (e.g., Romance vs.\ Indic recipients) are stable, the fine-grained
ordering of recipients is model-specific; exact values are reported in Appendix~\ref{app:ci}.

\paragraph{Dataset-Construction Sensitivity}
Since three datasets are machine-translated and \textit{Global-MMLU} is manually curated, we stratify \regMTXL{} by construction.
\textit{Global-MMLU} yields smaller gains and a higher harm rate, but target-language variance remains dominant in both subsets: 52.3\% for \textit{Global-MMLU} and 84.7\% for the machine-translated datasets. Donor rankings correlate moderately across subsets (Spearman $\rho \approx 0.46$), showing that construction and task content affect fine-grained rankings. High donor rankings in the machine-translated subset therefore do not necessarily carry over to the curated subset. Full results appear in Appendix~\ref{app:curated_vs_mt} \mbox{(Table~\ref{tab:mtcl_curated_vs_mt})}.

\paragraph{Adaptation-Setup Robustness}
In the rank ablation, average uplift increases through $r{=}64$ and then saturates, whereas harm rates rise steadily. We therefore choose $r{=}32$ for the main grid as an operating point near the gain--harm elbow: it improves average transfer over lower ranks while avoiding the substantially higher harm rates observed at $r{\geq}64$. On the same subset, per-cell deltas from LoRA and full fine-tuning are strongly correlated (Pearson $r \approx 0.83$) and differ by only $-0.04$~pp on average, although their effect directions agree in approximately $67\%$ of cells. Thus, the aggregate transfer structure is broadly preserved across adaptation settings, while individual cells can differ (Appendices~\ref{app:lora_rank} and \ref{app:fullft_vs_lora}).

\section{Discussion}
\label{sec:discussion}

Our results support three principles: (i) transfer success is driven primarily by target-side \emph{receptivity}; (ii) cross-task adaptation yields \emph{directed} rather than symmetric transfer; and (iii) transfer \emph{magnitudes} are broadly stable across model families, while fine-grained \emph{rankings} are not, so any robust guidance must remain coarse-grained.

\subsection{Asymmetric Donor--Recipient Structure}
The language donor--recipient landscape can be viewed as a directed transfer graph: languages differ in both how broadly they export useful updates and how readily they absorb incoming transfer. The target-dominated variance decomposition suggests that transfer is often receptivity-limited, since the same source update can yield markedly different gains across targets.
The negative association between syntactic distance and cross-lingual transfer further suggests that linguistic structure may partly shape how readily targets absorb incoming updates.
Task-level transfer also exhibits asymmetric donor--recipient roles: some tasks provide transferable signals that improve many others but are themselves easily disrupted, whereas others benefit from many sources yet provide little positive transfer in return.

\subsection{Stability: Where Regularities Hold and Where They Do Not}
Our stability analysis points to a distinction between \emph{coarse} regularities and \emph{fine-grained} predictability. The variance decomposition shows that transfer magnitudes are governed primarily by \emph{target-side receptivity}: some targets are systematically easier to improve across the models we study.
At the same time, low CI scores indicate that while coarse groupings of recipients into broad tiers (i.e., strong vs.\ weak recipient languages) are stable, the ordering within each tier is model-specific. Off-task regimes further amplify this uncertainty: cross-task transfer is driven more by idiosyncratic source--target interactions, making individual donor--recipient pairs harder to predict than in \regMTXL{} transfer.

\subsection{Practical Implications}
\label{sec:practical-implications}

These findings yield three practical guidelines for selecting fine-tuning sources and managing collateral risk.

\textbf{(1) Prioritise matched-task sources:} cross-language transfer is more reliable than cross-task transfer.

\textbf{(2) Condition donor choice on the recipient:} receptive targets permit broad choices, but fragile targets require an empirically tested shortlist; coarse tiers guide the search, while the best pair remains model-specific.

\textbf{(3) Monitor collateral degradation:} assess critical off-target capabilities, especially fragile tasks, not average gain alone.

Appendix~\ref{app:prompts_case_study} illustrates their application to \textit{Global-MMLU} in Bengali, considering both transfer gains and harm rates.

\section{Conclusion}
\label{sec:conclusion}

We presented a controlled analysis of cross-lingual and cross-task transfer under single-source LoRA fine-tuning across four benchmarks, eleven languages, and three open-weight LLM families at multiple scales. Although adaptation is net positive on average, pronounced structural asymmetries shape the transfer landscape.

In the matched-task cross-language regime, we find a stable hierarchy of language roles: high-resource Western European languages plus Chinese and Indonesian form a dense, high-transfer core, whereas low-resource Indic languages remain comparatively isolated recipients.

Task transfer is governed by complementary functional roles: \textit{Global-MMLU}, which assesses broad semantic knowledge, behaves as a near-universal recipient, whereas \textit{TruthfulQA}, which assesses truthfulness and factuality, is a strong donor but a fragile recipient. Mixed-effects variance decomposition shows that transfer success is dominated by target-side properties, especially in the \regMTXL{} regime, where the target language explains the majority of variance. In contrast, the fine-grained ordering of recipients for a given donor remains unstable across the models and scales we study.

\paragraph{Future directions.}
Future work should test whether these directed effects persist in larger models, additional architectures, and open-ended tasks evaluated by human annotators. A broader language inventory would also test whether the family-level hierarchy generalises beyond the present grid. Identifying measurable predictors of target receptivity could narrow the donor search without requiring an exhaustive transfer grid. Extending the framework beyond single-source PEFT to multilingual or multi-task mixtures would reveal whether the observed effects predict source interactions and whether mixtures can preserve gains for fragile recipients without compounding off-target harm.

\section*{Limitations}

Our conclusions are bounded by (i) model coverage (three open-weight families at $0.5$B--$8$B scales), (ii) the task suite and its multilingual construction, and (iii) a single adaptation regime (one-source PEFT with a fixed LoRA recipe).
Our language inventory is not balanced across families: four of the eleven languages are Romance, Germanic and Indic contain two languages each, and the remaining groups are each represented by a single language, limiting the generality of family-level comparisons.

Our evaluation is also limited to multiple-choice benchmarks. Although this format supports controlled automated evaluation, transfer dynamics may differ for open-ended generation.
Dataset construction may also affect the measured effects. Many evaluations rely on translation or post-editing, which can introduce artefacts favouring certain typologies, scripts, or registers. While this enables broad language coverage, our stratified comparison of curated vs.\ machine-translated benchmarks (Appendix~\ref{app:curated_vs_mt}) confirms that cross-lingual differences may partly reflect translation quality and editing practices in addition to intrinsic model transfer capabilities.
However, because \textit{Global-MMLU} is the only manually curated dataset, we cannot determine whether the observed differences reflect translation quality or task content.

Our evaluation protocol uses zero-shot prompting and fixed decoding. Few-shot or Chain-of-Thought (CoT) prompting and alternative decoding strategies may affect transfer patterns. Our primary metric is the absolute percentage-point change, and alternative metrics could alter the significance of the reported gains, particularly when baseline performance differs across targets. Finally, apart from a focused LoRA rank ablation on a small subset of models and sources (Appendix~\ref{app:lora_rank}), our experiments use a single LoRA recipe and examine single-source specialisation rather than multi-source or regularised schedules.

\section*{Acknowledgements}
\paragraph{Use of AI assistants.}
We used AI assistants for language polishing and minor code refactoring, alongside checks of structure, consistency, references, and formatting. All AI-assisted changes were reviewed and verified by the authors, who remain fully responsible for the manuscript, code, analyses, and conclusions.

\paragraph{Computational resources.}
The authors acknowledge the use of resources provided by the Isambard-AI National AI Research Resource (AIRR) \citep{mcintoshsmith2024isambardaileadershipclasssupercomputer}. Isambard-AI is operated by the University of Bristol and is funded by the UK Government’s Department for Science, Innovation and Technology (DSIT) via UK Research and Innovation; and the Science and Technology Facilities Council [ST/AIRR/I-A-I/1023].
Access was provided under the AIRR Early Access Project BYYG-VXG5-5.

\bibliography{custom}

\appendix

\section{Additional Experimental Details and Results}
\label{sec:appendix}

\subsection{Dataset Construction and Alignment}
\label{app:dataset_construction}
To ensure valid cross-lingual transfer comparisons, we aligned examples across languages and used disjoint train and test sets, preventing an example used for training in one language from appearing in another's test set. Training budgets are identical across languages within each dataset: ARC-Challenge, HellaSwag, and TruthfulQA provide 300 examples per language, while Global-MMLU provides all 215 available examples.

\paragraph{Alignment Algorithm}
For each benchmark, we identified the intersection of sample IDs across all available languages. From this common pool, we sampled fixed training and testing sets. This process is formalised in Algorithm~\ref{alg:alignment}.
\begin{algorithm}[ht!]
\SetAlgoLined
\KwIn{Languages $\mathcal{L}$, Source dataset $\mathcal{D}$,
      Train size $N_{\text{train}}$, Test size $N_{\text{test}}$,
      Seed $\sigma$}
\KwOut{Aligned training set $\mathcal{T}_{\text{train}}$,
       Evaluation set $\mathcal{T}_{\text{test}}$}
\KwData{Pseudo-random number generator \texttt{rng}}

\SetKwProg{Fn}{Function}{}{end}
\Fn{BuildCoreGrid($\mathcal{L}, \mathcal{D}$)}{
    $l_0 \leftarrow \mathcal{L}[0]$\;
    $\mathcal{I}_{\text{common}} \leftarrow \text{GetIDs}(\mathcal{D}, l_0)$\;
    \For{$l \in \mathcal{L} \setminus \{l_0\}$}{
        $\mathcal{I}_{l} \leftarrow \text{GetIDs}(\mathcal{D}, l)$\;
        $\mathcal{I}_{\text{common}} \leftarrow \mathcal{I}_{\text{common}} \cap \mathcal{I}_{l}$\;
    }

    \texttt{rng}.seed($\sigma$)\;
    Shuffle($\mathcal{I}_{\text{common}}$)\;

    $Ids_{\text{train}} \leftarrow \mathcal{I}_{\text{common}}[0 : N_{\text{train}}]$\;
    $Ids_{\text{test}} \leftarrow \mathcal{I}_{\text{common}}[N_{\text{train}} : N_{\text{train}} + N_{\text{test}}]$\;

    \For{$l \in \mathcal{L}$}{
        $\mathcal{T}_{\text{train}}[l] \leftarrow \text{Filter}(\mathcal{D}[l], Ids_{\text{train}})$\;
        $\mathcal{T}_{\text{test}}[l] \leftarrow \text{Filter}(\mathcal{D}[l], Ids_{\text{test}})$\;
    }
    \Return $\mathcal{T}_{\text{train}}, \mathcal{T}_{\text{test}}$\;
}
\caption{Cross-lingual data alignment. \texttt{rng} denotes a pseudo-random number generator used to shuffle the common ID pool.}
\label{alg:alignment}
\end{algorithm}
Table~\ref{tab:split_sources} summarises which released dataset splits act as sampling pools for each benchmark before applying the common-ID intersection and disjoint train/test sampling in Algorithm~\ref{alg:alignment}.
\begin{table}[t]
\centering
\small
\setlength{\tabcolsep}{3pt}
\renewcommand{\arraystretch}{1.05}
\begin{tabularx}{\columnwidth}{@{}l l l >{\raggedright\arraybackslash}X@{}}
\toprule
\textbf{Benchmark} &
\textbf{\shortstack[l]{Train pool}} &
\textbf{\shortstack[l]{Test pool}} &
\textbf{Align ID} \\
\midrule
ARC-Challenge    & train & test & id \\
Global-MMLU-Lite & dev   & test & sample\_id \\
HellaSwag        & val   & val  & id \\
TruthfulQA       & val   & val  & id (synthetic) \\
\bottomrule
\end{tabularx}
\caption{Released dataset splits used as sampling pools prior to the cross-lingual alignment procedure (Algorithm~\ref{alg:alignment}). For single-split benchmarks (HellaSwag, TruthfulQA), we sample disjoint train/test ID sets from the same released split.}
\label{tab:split_sources}
\end{table}

\subsection{Implementation Details}
\label{app:impl-details}
We load models with Hugging Face Transformers (v4.54.1) as \texttt{AutoModelForCausalLM} in BF16 and use each model's default tokeniser, configuring a padding token when needed. Text is tokenised with truncation and padding to fixed task-specific lengths.
Fine-tuning uses LoRA ($r{=}32$, $\alpha{=}64$) on attention projections (\texttt{q\_proj}, \texttt{k\_proj}, \texttt{v\_proj}, \texttt{o\_proj}) and MLP blocks (\texttt{gate\_proj}, \texttt{up\_proj}, \texttt{down\_proj}), optimising a standard causal language modelling objective over the concatenated ``prompt + gold answer''. For the multiple-choice benchmarks (\textit{ARC-Challenge}, \textit{Global-MMLU}, \textit{HellaSwag}), we supervise only the answer tokens by masking out the prompt in the loss; for \textit{TruthfulQA}, we supervise all non-padding tokens in the question--answer sequence. Training runs for 3 epochs in BF16 with AdamW (learning rate $5\times10^{-5}$, $\beta_1{=}0.9$, $\beta_2{=}0.999$, $\epsilon{=}10^{-8}$), a linear schedule with 10\% warmup, gradient clipping at 1.0, automatic batch-size discovery, and epoch-end evaluation and checkpoint saving.
All evaluations use the \emph{LM Evaluation Harness} v0.4.9.1 \citep{eval-harness} with fixed seeds:
\texttt{random\_seed=0}, \texttt{numpy\_seed=1234}, \texttt{torch\_seed=1234};
task-specific scoring/decoding follows harness defaults (e.g., log-likelihood for multiple-choice; otherwise greedy).

\paragraph{Compute Resources and Budget}
All experiments were run on a multi-GPU research cluster with NVIDIA GH200 GPUs (120\,GB each). The total compute budget across fine-tuning and evaluation was \(\approx 1400\) GPU-hours.

\subsection{LoRA Rank Ablation}
\label{app:lora_rank}

We perform a rank ablation on a representative subset of the transfer grid. Specifically, we vary the LoRA rank $r \in \{8,16,32,64,128\}$ for three distinct instruction-tuned models from different families and scales
(Llama-3.1-8B, Qwen2.5-1.5B, and Gemma-3-4B),
two fine-tuning tasks (\textit{ARC-Challenge} and \textit{Global-MMLU}), and three source languages
(Bengali, English, French), while keeping all other optimisation hyperparameters fixed.
For each rank, we set the LoRA scaling factor to $\alpha{=}2r$.
For each configuration and rank $r$, we compute the per-cell uplift $\Delta_r$ in percentage
points relative to the corresponding pre-adaptation score, using the same evaluation protocol as in the main analysis.

Table~\ref{tab:lora_rank_ablation} summarises mean uplift and harm rate across the ablation subset as a function of rank. Average uplift increases through $r{=}64$ and then saturates. Examining the distribution of per-cell uplifts, higher ranks primarily broaden the tails: the upper quantiles improve, but the lower quantiles become more negative. The harm rate rises steadily with rank, revealing a gain--harm trade-off rather than a uniformly optimal rank.

A regime-level breakdown (Table~\ref{tab:lora_rank_regimes}) shows that increasing the rank strengthens matched-task cross-language (\regMTXL{}) gains, but the largest ranks also produce higher harm rates and more extreme negative outliers. Off-task gains largely saturate around $r{=}32$--$64$, whereas off-task harm continues to rise.
At the level of individual transfer cells, per-cell deltas at $r{=}32$ and $r{=}64$ are highly correlated (Pearson $r(\Delta_{32}, \Delta_{64}) \approx 0.91$), indicating that increasing the rank from $32$ to $64$ largely rescales existing effects. Correlations with $r{=}128$ remain high but are noticeably weaker ($r \approx 0.71$). Together with the higher harm rate, this weaker correlation indicates less stable behaviour, with some donor--recipient pairs benefiting more strongly at the expense of others.
These results place $r{=}32$ near the elbow of the gain--harm trade-off: it improves average transfer over the smaller ranks while avoiding the sharp increase in harm observed at larger ranks.
We therefore fix $r{=}32$ in the main experiments.

\begin{table}[t!]
\centering
\small
\setlength{\tabcolsep}{4pt}%
\renewcommand{\arraystretch}{1.10}%
\begin{tabular}{lcc}
\toprule
\textbf{Rank $r$} & \textbf{Mean $\Delta$ (pp)} & \textbf{Harm rate (\%)} \\
\midrule
8   & 0.49 &  3.8 \\
16  & 0.64 &  5.2 \\
32  & 0.82 &  7.7 \\
64  & 0.93 & 13.0 \\
128 & 0.92 & 18.8 \\
\bottomrule
\end{tabular}
\caption{LoRA-rank ablation on three models, two source tasks, and three source languages. Uplift peaks at $r{=}64$, but harm rises steadily, placing $r{=}32$ near the gain--harm elbow.}
\label{tab:lora_rank_ablation}
\end{table}

\begin{table}[t!]
\centering
\small
\setlength{\tabcolsep}{4pt}%
\renewcommand{\arraystretch}{1.10}%
\begin{tabular}{llcc}
\toprule
\textbf{Rank $r$} & \textbf{Regime} &
\textbf{Mean $\Delta$ (pp)} &
\textbf{Harm rate (\%)} \\
\midrule
8   & \regMTXL{}    & 0.85 &  5.0 \\
8   & \regMTML{}    & 0.90 &  0.0 \\
8   & Off-task  & 0.34 &  3.5 \\
\addlinespace[2pt]
16  & \regMTXL{}    & 0.94 &  5.6 \\
16  & \regMTML{}    & 0.78 &  5.6 \\
16  & Off-task  & 0.50 &  5.1 \\
\addlinespace[2pt]
32  & \regMTXL{}    & 1.23 &  8.3 \\
32  & \regMTML{}    & 0.89 &  5.6 \\
32  & Off-task  & 0.68 &  7.6 \\
\addlinespace[2pt]
64  & \regMTXL{}    & 1.53 & 16.1 \\
64  & \regMTML{}    & 0.67 & 22.2 \\
64  & Off-task  & 0.82 & 11.8 \\
\addlinespace[2pt]
128 & \regMTXL{}    & 2.17 & 19.4 \\
128 & \regMTML{}    & 1.54 & 22.2 \\
128 & Off-task  & 0.67 & 18.5 \\
\bottomrule
\end{tabular}
\caption{Regime-level LoRA-rank ablation; off-task combines \regCTML{} and \regXTXL{}. Larger ranks strengthen \regMTXL{} but increase harm, while off-task gains plateau around $r{=}32$--$64$.}
\label{tab:lora_rank_regimes}
\end{table}

\subsection{Full Fine-Tuning vs.\ LoRA}
\label{app:fullft_vs_lora}

We performed a small-scale comparison between LoRA and full fine-tuning to confirm that our conclusions are not an artefact of using parameter-efficient adapters. We reused the ablation subset from Appendix~\ref{app:lora_rank}. We trained (i) LoRA models with $r{=}32$ (our main configuration) and (ii) fully fine-tuned models that update all parameters, using the same data splits and optimisation hyperparameters as in the main setup, but without adapters.
For each configuration, we computed per-cell uplifts $\Delta$ in percentage points relative to the corresponding pre-adaptation score and compared the LoRA and full fine-tuning deltas. The per-cell deltas are strongly correlated (Pearson $r \approx 0.83$), with approximately $67\%$ agreement in the direction of the effect. The average difference between LoRA and full fine-tuning is negligible (mean $\Delta_{\text{LoRA}} - \Delta_{\text{full}} \approx -0.04$~pp, median $0$~pp), indicating that LoRA provides an essentially unbiased approximation to full fine-tuning while preserving the overall structure of transfer effects, even though individual cells exhibit some local noise.

\subsection{Curated vs.\ Machine-Translated Datasets}
\label{app:curated_vs_mt}

Table~\ref{tab:mtcl_curated_vs_mt} summarises the outcomes and subset-specific
decompositions. The target remains the largest variance component in both
subsets and is especially dominant among the machine-translated datasets.
The \textit{Global-MMLU} subset has larger model and source components,
signalling greater model and donor sensitivity. Donor scores correlate only
moderately (Spearman $\rho \approx 0.46$), showing that construction and task
content affect fine-grained patterns without explaining away target-side
dominance.

\subsection{Stability of Transfer Rankings}
\label{app:ci}

While the variance decomposition highlights strong structural drivers of
transfer magnitude, we also assess the stability of recipient
\textit{rankings} across different models. We define the
Consistency Index to quantify whether different models agree on which targets benefit most from a given donor. Formally, for each source $s$ (a specific task or language) and transfer regime, we collect for every model $m$ the vector of transfer effects
across recipients:
\[
\boldsymbol{\Delta}^{(s)}_{m}
= \bigl(\Delta_{m,s\to r}\bigr)_{r\in\mathcal{R}_{s}},
\]
where $\mathcal{R}_{s}$ is the set of valid recipients for source $s$ in
that regime. We define the per-source Consistency Index as the mean
pairwise Kendall rank correlation coefficient ($\tau$) between models:
\[
\mathrm{CI}(s)
\;=\;
\frac{\sum_{i<j}
  \tau\!\bigl(\boldsymbol{\Delta}^{(s)}_{m_i},
              \boldsymbol{\Delta}^{(s)}_{m_j}\bigr)}
     {\binom{M}{2}},
\]
where $M$ is the number of models. A high CI implies that different models
agree on the ordering of best-to-worst recipients for a given donor.
Table~\ref{tab:consistency-index} shows that, despite the target dimension
explaining the majority of variance in magnitudes, the ranking consistency
is universally low. In the \regMTXL{} regime, for instance, while all models
agree on broad tiers (e.g., Romance languages consistently outperform Indic
languages as recipients), the fine-grained ordering within these tiers is
highly idiosyncratic. The specific permutation of close neighbours (e.g.,
ranking Spanish vs.\ Italian vs.\ Portuguese) appears to be driven by
model-specific factors rather than universal linguistic properties.
Consequently, while one can predict that a high-resource language group
will benefit from transfer, predicting the \textit{single best} recipient
language for a specific model remains difficult without empirical testing.

\begin{table}[t!]
\centering
{
\begin{tabular}{lcc}
\toprule
\textbf{Regime} & \textbf{Median $\tau$} & \textbf{IQR} \\
\midrule
\textbf{\regMTXL} & 0.03 & $[0.00, 0.08]$ \\
\textbf{\regCTML} & 0.11 & $[-0.04, 0.32]$ \\
\textbf{\regXTXL} & 0.13 & $[0.05, 0.18]$ \\
\bottomrule
\end{tabular}
}
\caption{Median Consistency Index (Kendall $\tau$) and interquartile range (IQR) across sources for each transfer regime. The low median values indicate weak cross-model agreement in fine-grained recipient rankings.}
\label{tab:consistency-index}
\end{table}

\begin{table}[t!]
\centering
{
\setlength{\tabcolsep}{4pt}%
\renewcommand{\arraystretch}{1.0}%
\begin{tabular}{lcc}
\toprule
\textbf{Task (Benchmark)} & \textbf{Donor} & \textbf{Recipient} \\
\midrule
ARC-Challenge  & 1.11 &  0.03 \\
Global-MMLU    & 0.03 &  2.78 \\
HellaSwag      & 0.92 &  0.52 \\
TruthfulQA     & 1.17 & -0.10 \\
\bottomrule
\end{tabular}
}
\caption{Task donor and recipient scores in \regCTML{}, averaged across languages, model families, and scales. \textit{TruthfulQA} is the strongest donor but weakest recipient; \textit{Global-MMLU} shows the reverse pattern.}
\label{tab:task-donor-recipient}
\end{table}

\begin{table}[t!]
\centering
{\small
\setlength{\tabcolsep}{4pt}%
\renewcommand{\arraystretch}{1.0}%
\begin{tabular}{lrr}
\toprule
\textbf{Language} & \textbf{Donor} & \textbf{Recipient} \\
\midrule
ar & 0.99 & 1.14 \\
bn & 0.97 & 0.60 \\
de & 1.27 & 1.46 \\
en & 1.53 & 1.30 \\
es & 1.18 & 2.13 \\
fr & 1.47 & 1.05 \\
hi & 0.98 & 0.84 \\
id & 1.33 & 1.11 \\
it & 1.24 & 1.69 \\
pt & 1.50 & 1.21 \\
zh & 1.25 & 1.19 \\
\bottomrule
\end{tabular}
}
\caption{Language donor and recipient scores in \regMTXL{}, averaged across tasks, model families, and scales. English is the strongest donor, Spanish the strongest recipient, and Bengali the weakest recipient.}
\label{tab:lang_donor_recipient}
\end{table}

\subsection{Linguistic Distance Correlations}
\label{app:ling_corr}

We conducted a correlational analysis to test whether transfer performance is associated with three types of typological similarity.
We focused on the \regMTXL~regime to isolate linguistic effects from task transfer effects. For every pair of distinct source and target languages, we computed the Spearman rank correlation ($\rho$) between the observed Transfer Score ($\Delta$) and three standard linguistic distance metrics provided by the URIEL knowledge base \citep{littell-etal-2017-uriel}:

\begin{enumerate}
    \item \textbf{Syntactic Distance:} Cosine distance between KNN-predicted syntactic feature vectors, including word-order features.
    \item \textbf{Phonological Distance:} Cosine distance between KNN-predicted phonological feature vectors.
    \item \textbf{Inventory Distance:} Cosine distance between KNN-predicted phonological segment-inventory feature vectors.
\end{enumerate}

As shown in Table~\ref{tab:ling_corr_appendix}, syntactic distance has the strongest negative association with transfer: language pairs with more similar syntactic feature vectors tend to exhibit larger transfer gains. Phonological distance has a weaker but statistically significant negative association, whereas inventory distance is not statistically significant. Thus, among the three URIEL distance metrics considered, syntactic distance is the strongest correlate of transfer. These results do not measure lexical overlap or establish a causal mechanism.

\begin{figure*}[b!]
  \centering
  \includegraphics[width=0.85\textwidth]{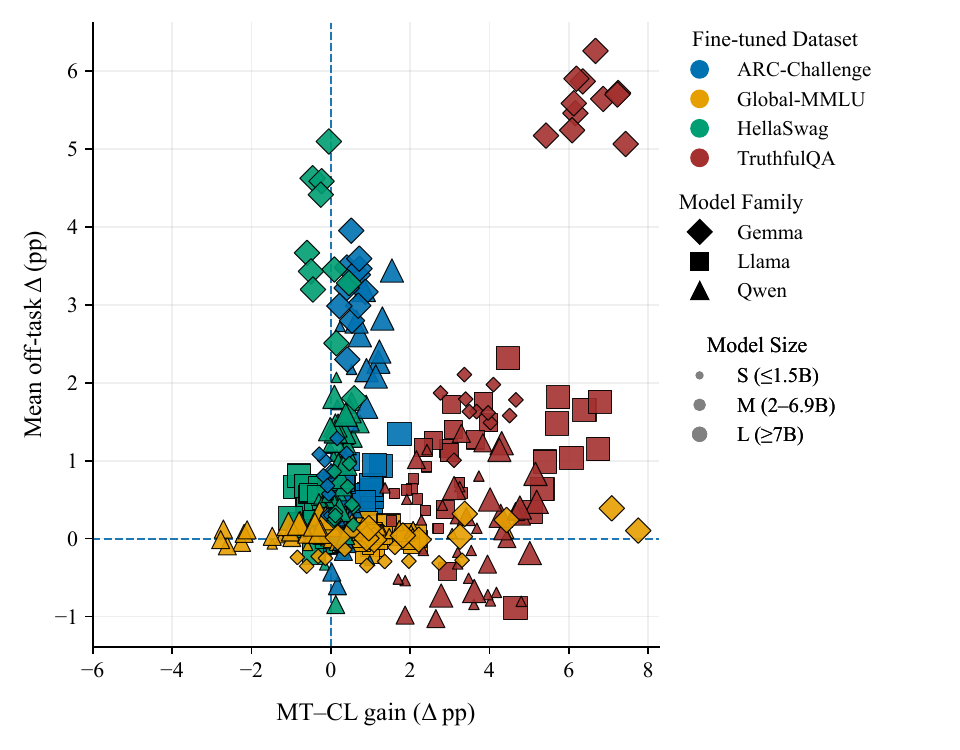}
  \caption{Complete configuration-level matched-task and off-task transfer map. Each point represents one model--source-cell configuration, aggregated over the three experiment seeds. Colour identifies the source task, marker shape the model family, and marker size the size bucket ($S$: $\leq 1.5$B; $M$: 2--6.9B; $L$: $\geq 7$B).}
  \label{fig:pareto-onoff-full}
\end{figure*}

\subsection{Prompt Templates and Qualitative Case Study}
\label{app:prompts_case_study}

\paragraph{Prompt Templates}
For fine-tuning, we convert each translated benchmark example into a short, language-parallel prompt. Across all languages, the question stem and answer options are in the target language, while a small set of control tokens (e.g.\ \texttt{Question}, \texttt{Answer}, option labels) are kept fixed.

\medskip
\noindent\textbf{ARC-Challenge.}
Each example is formatted as a multiple-choice question with four options:
\begin{quote}\small\ttfamily
Question: <question>\\
A) <option\_A>\\
B) <option\_B>\\
C) <option\_C>\\
D) <option\_D>\\
Answer:
\end{quote}

\medskip
\noindent\textbf{Global-MMLU-Lite.}
We use a compact multiple-choice template without an explicit \texttt{Question:} prefix:
\begin{quote}\small\ttfamily
<question>\\
A. <option\_A>\\
B. <option\_B>\\
C. <option\_C>\\
D. <option\_D>\\
Answer:
\end{quote}

\medskip
\noindent\textbf{HellaSwag.}
For \textit{HellaSwag}, we first concatenate the activity label and context into a single description, then present four possible endings:
\begin{quote}\small\ttfamily
<activity\_label>: <context>\\
A) <ending\_1>\\
B) <ending\_2>\\
C) <ending\_3>\\
D) <ending\_4>\\
Answer:
\end{quote}
The model is trained to generate the letter corresponding to the correct continuation.

\medskip
\noindent\textbf{TruthfulQA.}
For \textit{TruthfulQA} (MC1), we collapse the multiple-choice format into a short answer prompt and train the model to produce the gold answer text:
\begin{quote}\small\ttfamily
Question: <question>\\
Answer: <correct\_answer>
\end{quote}
Here the supervision covers the full gold answer string (in the target language), rather than an option label.

\paragraph{Working Practitioner Example}
As a concrete illustration, consider a practitioner who wishes to improve Bengali performance on \textit{Global-MMLU} without harming other capabilities. They first inspect the language-level donor and recipient scores in Table~\ref{tab:lang_donor_recipient}, noting that Bengali is a relatively weak recipient compared with high-resource languages such as English, Italian, or Spanish. This suggests that Bengali tends to benefit less from others and is therefore a harder target. Next, they consult the \regMTXL{} language-to-language transfer matrix in Table~\ref{tab:lang_to_lang_transfer} and focus on the Bengali column, where English, Chinese, and Hindi provide the strongest average incoming gains. They consider these pair-specific gains alongside the global donor scores in Table~\ref{tab:lang_donor_recipient} to form a shortlist of candidate donors. Finally, the practitioner uses the \regMTXL{} breakdown in Table~\ref{tab:mtcl_harm_breakdown} to assess Bengali's general recipient and harm profile, and consults the matched-task versus off-task profiles in Figure~\ref{fig:pareto_onoff} to identify model- and task-level settings with negative off-task effects. These aggregate analyses yield a shortlist of candidate donor languages. Each candidate should then be validated with the intended model on the target dataset, while monitoring critical off-target capabilities for degradation.

\paragraph{Resource Access and Licensing}
\label{app:licenses}
Our \href{https://github.com/kdymkiewicz/Donors-and-Recipients}{public project repository} provides the deterministic split manifests and per-cell main-grid results described in this paper. The four language-aligned datasets are available directly from Hugging Face: \href{https://huggingface.co/datasets/Dr4kl3s/arc_challenge_core_grid_seed42}{ARC-Challenge}, \href{https://huggingface.co/datasets/Dr4kl3s/global_mmlu_lite_core_grid_seed42}{Global-MMLU-Lite}, \href{https://huggingface.co/datasets/Dr4kl3s/hellaswag_coregrid_seed42}{HellaSwag}, and \href{https://huggingface.co/datasets/Dr4kl3s/truthfulqa_coregrid_seed42}{TruthfulQA}. The code is released under Apache 2.0, and the numerical result artefact under CC BY 4.0. The aligned datasets remain governed by their respective upstream licences and access conditions; neither licence applies to third-party data. We do not redistribute the starting instruction-tuned checkpoints, fine-tuned weights, or adapters.

\begin{table*}[t!]
  \centering
  \small
  \setlength{\tabcolsep}{5pt}
  \begin{tabular}{llrrrr}
    \toprule
    Group & Key & Mean $\Delta$ (pp) & Median $\Delta$ (pp) & Positive-transfer rate (\%) & Harm rate (\%) \\
    \midrule
    Dataset & \textit{ARC-Challenge}      &  0.41 &  0.33 & 66.5 &  2.3 \\
    Dataset & \textit{Global-MMLU-Lite}   &  0.69 &  0.42 & 58.7 & 21.5 \\
    Dataset & \textit{HellaSwag}          & -0.03 & -0.08 & 43.7 &  4.5 \\
    Dataset & \textit{TruthfulQA}         &  3.91 &  3.75 & 96.8 &  0.1 \\
    \midrule
    Family  & Gemma                       &  1.86 &  0.75 & 71.8 &  8.6 \\
    Family  & Llama                       &  1.17 &  0.50 & 64.4 &  3.9 \\
    Family  & Qwen                        &  1.00 &  0.42 & 65.2 &  8.8 \\
    \midrule
    Language & ar                         &  1.14 &  0.58 & 67.8 &  4.7 \\
    Language & bn                         &  0.60 &  0.25 & 63.9 &  3.1 \\
    Language & de                         &  1.45 &  0.50 & 68.9 &  7.2 \\
    Language & en                         &  1.30 &  0.71 & 71.4 &  4.7 \\
    Language & es                         &  2.13 &  0.71 & 65.3 &  8.9 \\
    Language & fr                         &  1.05 &  0.42 & 60.0 & 11.1 \\
    Language & hi                         &  0.83 &  0.33 & 60.8 & 10.3 \\
    Language & id                         &  1.11 &  0.38 & 64.7 &  9.2 \\
    Language & it                         &  1.69 &  0.75 & 74.2 &  3.6 \\
    Language & pt                         &  1.21 &  0.42 & 64.7 &  7.2 \\
    Language & zh                         &  1.19 &  0.92 & 68.9 &  8.3 \\
    \midrule
    Size bucket & L ($\geq$7B)            &  1.54 &  0.83 & 69.8 &  6.6 \\
    Size bucket & M (2--6.9B)             &  1.38 &  0.50 & 64.3 &  9.8 \\
    Size bucket & S ($\leq$1.5B)          &  1.00 &  0.42 & 66.3 &  5.3 \\
    \midrule
    Overall & All \regMTXL{}                  &  1.25 &  0.50 & 66.4 &  7.1 \\
    \bottomrule
  \end{tabular}
  \caption{\regMTXL{} outcomes by target dataset, model family, language, and size bucket. Values are in percentage points; harm denotes $\Delta \leq -1.0$~pp. \textit{TruthfulQA} has the largest uplift, whereas \textit{Global-MMLU} has the highest harm rate.}
  \label{tab:mtcl_harm_breakdown}
\end{table*}

\begin{table*}[t!]
\centering
\small
\setlength{\tabcolsep}{6pt}%
\begin{tabular}{lrrrrrr}
\toprule
& \multicolumn{2}{c}{\textbf{Transfer outcomes}}
& \multicolumn{4}{c}{\textbf{REML variance share (\%)}} \\
\cmidrule(lr){2-3}\cmidrule(lr){4-7}
\textbf{Subset} & \textbf{Mean $\Delta$ (pp)} & \textbf{Harm rate (\%)}
& \textbf{Model} & \textbf{Source} & \textbf{Target} & \textbf{Residual} \\
\midrule
Curated (Global-MMLU) & 0.69 & 21.5 & 16.6 & 14.2 & 52.3 & 16.9 \\
Machine-translated (three datasets) & 1.43 & 2.3 & 2.7 & 4.5 & 84.7 & 8.1 \\
All datasets & 1.25 & 7.1 & 4.2 & 8.8 & 75.4 & 11.6 \\
\bottomrule
\end{tabular}
\caption{\regMTXL{} outcomes and REML variance shares for curated \textit{Global-MMLU} versus the three machine-translated datasets. Harm denotes $\Delta \leq -1.0$~pp. Target variation dominates both subsets, while the curated subset has lower uplift and substantially higher harm.}
\label{tab:mtcl_curated_vs_mt}
\end{table*}

\begin{table*}[t!]
\centering
{\small
\setlength{\tabcolsep}{3pt}%
\renewcommand{\arraystretch}{1.10}%
\begin{tabular}{lrrrr}
\toprule
\textbf{Fine-tuned task} & \textbf{ARC-Challenge} & \textbf{TruthfulQA} & \textbf{HellaSwag} & \textbf{Global-MMLU} \\
\midrule
ARC-Challenge & --    & -0.01 & 0.24 & 3.09 \\
TruthfulQA    & -0.43 & --    & 1.24 & 2.71 \\
HellaSwag     & 0.36  & -0.15 & --   & 2.55 \\
Global-MMLU   & 0.14  & -0.14 & 0.09 & --   \\
\bottomrule
\end{tabular}
}
\caption{Task-to-task transfer matrix in the \regCTML{} regime. Rows are fine-tuned tasks; columns are evaluated tasks. Diagonal entries (--) are same-task and excluded. Values are $\Delta$ in percentage points, averaged across languages, model families, and scales.}
\label{tab:task_to_task_transfer}
\end{table*}

\begin{table*}[t!]
    \centering
    \small
    \renewcommand{\arraystretch}{1.2}
    \begin{tabular}{lccc}
    \toprule
    \textbf{Linguistic Metric} & \textbf{Spearman $\rho$} & \textbf{$p$-value} & \textbf{Significance} \\
    \midrule
    Syntactic Distance    & $\mathbf{-0.605}$ & $\mathbf{<0.001}$ & *** \\
    Phonological Distance & $-0.206$ & $0.031$  & * \\
    Inventory Distance    & $-0.157$ & $0.102$  & n.s. \\
    \bottomrule
    \end{tabular}
    \caption{Spearman correlations between linguistic distance and \regMTXL{} transfer; *** denotes $p<0.001$, * denotes $p<0.05$, and n.s. denotes non-significance. Syntactic distance has the strongest negative association; inventory distance is not significant.}
    \label{tab:ling_corr_appendix}
\end{table*}

\begin{table*}[t!]
\centering
{\small
\setlength{\tabcolsep}{3pt}%
\renewcommand{\arraystretch}{1.05}%
\begin{tabular*}{\textwidth}{@{\extracolsep{\fill}}lrrrrrrrrrrr}
\toprule
 & \textbf{ar} & \textbf{bn} & \textbf{de} & \textbf{en} & \textbf{es} & \textbf{fr} & \textbf{hi} & \textbf{id} & \textbf{it} & \textbf{pt} & \textbf{zh} \\
\midrule
ar & --    & 0.55 & 1.26 & 1.00 & 1.60 & 0.84 & 0.60 & 0.68 & 1.31 & 1.11 & 0.96 \\
bn & 0.90 & --    & 1.28 & 0.71 & 1.46 & 0.85 & 1.09 & 0.61 & 1.08 & 0.74 & 0.97 \\
de & 1.19 & 0.56 & --    & 1.57 & 2.10 & 1.10 & 0.91 & 1.27 & 1.83 & 1.00 & 1.14 \\
en & 1.15 & 0.91 & 1.61 & --    & 2.41 & 1.61 & 0.95 & 1.47 & 1.69 & 1.74 & 1.78 \\
es & 1.01 & 0.60 & 1.49 & 1.35 & --    & 0.87 & 0.67 & 1.21 & 1.99 & 1.50 & 1.10 \\
fr & 1.19 & 0.59 & 1.67 & 1.60 & 2.45 & --    & 0.81 & 1.48 & 2.19 & 1.48 & 1.26 \\
hi & 0.95 & 0.67 & 1.37 & 0.90 & 1.44 & 0.84 & --    & 0.66 & 1.31 & 0.80 & 0.85 \\
id & 1.16 & 0.55 & 1.58 & 1.52 & 2.05 & 1.20 & 0.80 & --    & 2.00 & 1.20 & 1.24 \\
it & 1.27 & 0.28 & 1.48 & 1.25 & 2.90 & 0.85 & 0.77 & 1.09 & --    & 1.26 & 1.21 \\
pt & 1.28 & 0.57 & 1.60 & 1.80 & 2.99 & 1.40 & 0.73 & 1.36 & 1.89 & --    & 1.40 \\
zh & 1.31 & 0.72 & 1.19 & 1.34 & 1.87 & 0.96 & 1.01 & 1.27 & 1.58 & 1.25 & --    \\
\bottomrule
\end{tabular*}
}
\caption{Language-to-language transfer matrix in the \regMTXL{} regime. Rows are fine-tuned (source) languages; columns are evaluated (target) languages. Diagonal entries (--) denote same-language fine-tuning and are excluded. Values are $\Delta$ in percentage points, averaged across tasks, model families, and scales.}
\label{tab:lang_to_lang_transfer}
\end{table*}

\clearpage

\begin{sidewaystable*}
\centering
\begin{adjustbox}{max width=\textheight, max totalheight=\textwidth, keepaspectratio}
{\small
\setlength{\tabcolsep}{3pt}%
\renewcommand{\arraystretch}{1.10}%
\begin{tabularx}{\textheight}{l l l l L}
\toprule
\textbf{Dataset} & \textbf{Task type} & \textbf{Construction type} &
\textbf{Training / Validation / Test ($N$)} &
\textbf{Short description / rationale} \\
\midrule
\textit{ARC-Challenge} & Science QA (commonsense + background) &
\textbf{Machine-translated} (LLM; parallelised) &
270 / 30 / 400 &
Knowledge-intensive multiple-choice reasoning requiring background/world knowledge beyond surface cues; used to probe cross-lingual transfer for reasoning that relies on external knowledge under parallelised content. \\
\addlinespace[2pt]
\textit{TruthfulQA} & Factuality / truthful QA &
\textbf{Machine-translated} (LLM; parallelised) &
270 / 30 / 400 &
Stress-tests truthfulness against common misconceptions and misleading prompts; included to examine whether gains in other tasks/languages spill over or harm factual reliability when content is held constant across languages. \\
\addlinespace[2pt]
\textit{HellaSwag} & Commonsense inference &
\textbf{Machine-translated} (LLM; parallelised) &
270 / 30 / 400 &
Adversarial commonsense multiple-choice questions designed to reduce annotation artefacts; used to test everyday reasoning transfer across scripts and typologies under matched scenarios. \\
\addlinespace[2pt]
\textit{Global-MMLU-Lite} & Knowledge-intensive reasoning &
\textbf{Human-translated / curated} &
193 / 22 / 400 &
Broad subject-knowledge dataset; we use the Global-MMLU-Lite subset with carefully curated multilingual data, providing a complementary signal to the machine-translated datasets. \\
\bottomrule
\end{tabularx}
}
\end{adjustbox}
\caption{Datasets. We report task types, construction methods, and sample sizes for the custom \textbf{Training}/\textbf{Validation}/\textbf{Test} splits; \textbf{Val} is 10\% of the original training split (rounded to whole examples) used for early stopping. All datasets cover the same 11 languages, enabling parallel training and evaluation across the transfer matrix.}
\label{tab:benchmarks-languages}
\end{sidewaystable*}

\end{document}